\documentclass[conference]{IEEEtran}
\IEEEoverridecommandlockouts
\usepackage{cite}
\usepackage{amsmath,amssymb,amsfonts}
\usepackage{algorithmic}
\usepackage{graphicx}
\usepackage{textcomp}
\usepackage{xcolor}
\usepackage{booktabs} 
\usepackage{url} 
\usepackage{hyperref} 
\hypersetup{colorlinks=false, pdfborder={0 0 0}}
\usepackage{cleveref} 
\usepackage{rotating} 
\def\BibTeX{{\rm B\kern-.05em{\sc i\kern-.025em b}\kern-.08em
    T\kern-.1667em\lower.7ex\hbox{E}\kern-.125emX}}

\begin{document}

\title{A Text-to-tabular Approach to Generate Synthetic Patient Data using LLMs
\thanks{Accepted to the 13th IEEE International Conference on Healthcare Informatics (IEEE
ICHI 2025), Rende (CS), Calabria, Italy,}
}
\author{\IEEEauthorblockN{Margaux T\"ornqvist}
\IEEEauthorblockA{\textit{Quinten Health}\\
Paris, France \\
m.tornqvist@quinten-health.com}
\and
\IEEEauthorblockN{Jean-Daniel Zucker}
\IEEEauthorblockA{\textit{IRD, Sorbonne Université, UMMISCO}\\
\textit{SU, INSERM, NUTRIOMICS,} Paris, France \\
jean-daniel.zucker@ird.fr}
\and
\IEEEauthorblockN{Tristan Fauvel}
\IEEEauthorblockA{\textit{Quinten Health}\\
Paris, France \\
t.fauvel@quinten-health.com}
\and
\IEEEauthorblockN{Nicolas Lambert}
\IEEEauthorblockA{\textit{Quinten Health}\\
Paris, France \\
n.lambert@quinten-health.com}
\and
\IEEEauthorblockN{Mathilde Berthelot}
\IEEEauthorblockA{\textit{Quinten Health}\\
Paris, France \\
m.berthelot@quinten-health.com}
\and
\IEEEauthorblockN{Antoine Movschin}
\IEEEauthorblockA{\textit{Quinten Health}\\
Paris, France \\
a.movschin@quinten-health.com}
}

\maketitle

\begin{abstract}
Access to large-scale high-quality healthcare databases is key to accelerate medical research and make insightful discoveries about diseases. However, access to such data is often limited by patient privacy concerns, data sharing restrictions and high costs. To overcome these limitations, synthetic patient data has emerged as an alternative. However, synthetic data generation (SDG) methods typically rely on machine learning (ML) models trained on original data, leading to the data scarcity problem. We propose an approach to generate synthetic tabular patient data that does not require access to the original data, but only a description of the desired database. We harness prior medical knowledge and the in-context learning capabilities of large language models (LLMs) to perform zero-shot generation of realistic patient data, even in low-resource settings. We quantitatively evaluate our approach against state-of-the-art SDG Models, using fidelity, privacy, and utility metrics. Our results show that while LLMs may not match the performance of state-of-the-art models trained on the original data, they effectively generate realistic patient data with well-preserved clinical correlations. An ablation study highlights key elements of our prompt that contribute to the generation of high-quality synthetic patient data. This approach, which is easy to use and does not require original data or advanced ML skills, is particularly valuable for quickly generating custom-designed patient data, supporting project implementation, and providing educational resources.
\end{abstract}


This study uses data from the Parkinson’s Progression Markers Initiative (PPMI) database (RRID:SCR 006431) and from the Alzheimer’s Disease Neuroimaging Initiative (ADNI) database (RRID:SCR 003007). For up-to-date information on these studies, visit www.ppmi-info.org and www.adni-info.org. The code used in this study and Supplementary Materials are available at \url{https://github.com/quinten-health-os/synth-data-gen-from-text}.

\section{Introduction}
Accelerating medical research using machine learning (ML) requires the availability of large, comprehensive, and complete patient data. In fact, to extract clinically relevant insights, ML models typically require access to extensive datasets that accurately represent the target population, maintain balanced class distributions, and without too many missing values. However, concerns regarding patient privacy, cost, and availability often limit access to such valuable data. Synthetic data generation (SDG) methods that aim at generating data that mimic real world data (RWD) have emerged as a promising solution to bypass these limits \cite{gonzales_synthetic_2023}. Indeed, synthetic data which maintain statistical properties of the original data while preserving patient privacy, can be used for a variety of analyzes to accelerate medical research. These include privacy-preserved data sharing \cite{yale_generation_2020, hernandez_synthetic_2022, eckardt_mimicking_2023}, data augmentation \cite{perez_effectiveness_2017, seedat_curated_2024, nakada_synthetic_2024} and cohort enrichment \cite{pauley_t1dctegui_2023}. In addition, SDG could accelerate and reduce the costs of randomized clinical trials (RCTs) by enhancing or replacing control arms with digital twins \cite{lhostis_knowledge-based_2023, alam_digital_2024}. 

In this study, we present an innovative method for generating synthetic patient data using large language models (LLMs). Since LLMs are partially trained on biomedical corpora—including publicly available abstracts from databases like PubMed, we hypothesize that these models could extract meaningful relationships between diseases, clinical symptoms, and patient characteristics, which is certainly key for generating high-quality synthetic patient data. As originally proposed by \cite{borisov_language_2023}, we use a pre-trained LLM directly as the SDG Model. However, unlike previous tabular-to-tabular approaches relying on a learning phase on the original data and a generation phase of synthetic tabular data, we introduce a novel text-to-tabular approach requiring no original data and no fine-tuning of the model. This method involves generating synthetic tabular data directly from a textual description of the desired database. The main motivation is to circumvent the need for inaccessible original patient-level data, and the high computational cost of training or fine-tuning SDG Models. 

We evaluate our approach by generating synthetic clinical characteristics of patients with Parkinson’s Disease (PD) from the PPMI Clinical database and of patients with Alzheimer's Disease (AD) from the ADNI database. We conduct a rigorous assessment of the synthetic data, evaluating fidelity  — how accurately the synthetic data mimics the original data, utility — how effectively the synthetic data enhances performance in the target task, and privacy, with both a qualitative and a quantitative assessment using different metrics. Moreover, we conduct a comparative analysis of our text-to-tabular approach with state-of-the-art SDG tabular-to-tabular methods to highlight the advantages and limitations of using LLMs with no original data. Finally, we present an ablation study aiming to highlight the features of our prompt that contribute to the generation of high-quality synthetic data. Of note, our ultimate goal of generating such synthetic data is for research purposes and post hoc analyses.

In this work, we present a framework for generating synthetic tabular patient data. Our contributions are as follows.
\begin{itemize}
    \item We propose a text-to-tabular SDG approach using a frozen LLM (pre-trained but not fine-tuned) as the SDG Model conditioned solely on a database description. Unlike most generative models for tabular data, our approach does not require access to the original data. This makes it particularly useful in situations where the original data is difficult or timely to obtain or subject to privacy concerns. 
    \item We conduct a comprehensive evaluation of our framework against state-of-the-art tabular-to-tabular SDG Models and show that it performs competitively, despite not requiring access to the original data. Our results demonstrate that the synthetic patient data generated by LLMs has the potential to replicate statistical properties and patterns found in real-world patient data.
\end{itemize}

\section{Related Works}
\label{sec:related works}

Following advances in generative AI for images and text, attention has recently shifted towards tabular data. Various tabular-to-tabular SDG methods, involving a training phase of a model on the original data, have shown promising results in generating realistic RWD \cite{borisov_deep_2024}.

\paragraph{Probabilistic models} SDG can be effectively achieved using probabilistic graphical models such as Bayesian networks \cite{young_using_2009, zhang_privbayes_nodate} and Gaussian Copulas (GCs), which model the joint distribution of variables from their marginal distributions \cite{patki_synthetic_2016, kamthe_copula_2021}.

\paragraph{Generative adversarial networks (GANs)} have demonstrated remarkable capabilities in image generation \cite{goodfellow_generative_2014}, and have been successfully extended to tabular data \cite{borisov_deep_2024}. Among recent advances, CTGAN stands out for its effectiveness in synthesizing mixed-type data, that is, data with a mix of continuous and discrete variables, a long-standing challenge in tabular data generation \cite{xu_modeling_2019}. CTGAN introduced a conditional vector in the generation process to handle unbalanced datasets and mode-specific normalization to effectively represent distributions with multiple modes. Building on this, \cite{zhao_ctab-gan_2021} introduced CTAB-GAN, which incorporates additional downstream losses and a novel encoding technique for the conditional vector, allowing efficient modeling of unbalanced or skewed data and mixed-type variables. Other tabular GANs, such as DP-GAN \cite{xie_differentially_2018}, Pate-GAN \cite{jordon_pate-gan_2018} and CTAB-GAN+ \cite{zhao_ctab-gan_2022}, also introduced specific losses and training techniques to ensure the privacy of the original data.

\paragraph{Variational autoencoders (VAEs) and diffusion models} Alongside CTGAN, \cite{xu_modeling_2019} proposed TVAE, which outperformed their suggested GAN approach. Diffusion models, which have gained significant attention in the computer vision domain \cite{austin_structured_2021}, have also inspired approaches to generate tabular data. Notable examples include TABDDPM, which has shown remarkable performance in modeling heterogeneous features \cite{kotelnikov_tabddpm_2022}, and diffusion flow-based gradient boosted trees \cite{jolicoeur-martineau_generating_2024}.

\paragraph{LLMs} The application of LLMs, which have demonstrated significant proficiency in the textual domain, has inspired many to explore their use in the generation of realistic synthetic tabular data. The results obtained have been unexpectedly promising, outperforming state-of-the-art GANs in benchmark tabular datasets \cite{borisov_language_2023}, indicating the potential of LLMs in this area. Currently, most SDG approaches that use LLMs involve fine-tuning the models on the original data or custom prompts that include the original data.
\cite{borisov_language_2023} pioneered the use of LLMs for the SDG of tabular data, introducing GReaT, a method that uses pre-trained LLMs and that significantly outperforms GANs in synthesizing high-quality tabular data. Building on this foundation, several LLM-based SDG Models with new features were developed: ReaLTAbFormer \cite{solatorio_realtabformer_2023}, TabuLa \cite{zhao_tabula_2023} and DP-LLMTGen \cite{tran_differentially_2024}. Additionally, \cite{seedat_curated_2024} proposed CLLM, a curated LLM for data augmentation in small datasets, utilizing a frozen LLM and an add-on curation filter to generate high-quality and diverse new samples using the original data. In the latest research has been proposed MALLM-GAN, a framework inspired by the architecture of GANs using a frozen LLM within both the generator and optimizer to produce synthetic data \cite{ling_mallm-gan_2024}.  Moreover, \cite{xu_are_2024} showed that the random order permutation of variables during fine-tuning, as proposed by \cite{borisov_language_2023}, does not allow an accurate estimation of the conditional mixture of distributions and that permutations corresponding to learned functional dependencies between variables are more appropriate.

All the works cited above require access to the original data. A notable work introducing a LLM that does not require access to some is LLM-VP, a virtual patient generator that refines \textit{prompts} through iterative interaction with an LLM and clinicians \cite{cook_creating_2024}. However, although relevant to SDG, it is primarily designed to simulate real-life clinical scenarios for educational purposes, rather than for SDG in scenarios where patient data is explicitly unavailable.

\section{Materials and Methods}
\label{sec:methods}
\subsection{Data}

\subsubsection{PPMI}
Parkinson’s disease (PD) is a complex, heterogeneous, age-related neurodegenerative disease that is often described by a combination of core diagnostic symptoms, including bradykinesia, rigidity, tremor and postural instability \cite{politis_parkinsons_2010}. PPMI is a global observational study to establish biomarker-defined cohorts and identify clinical, imaging, genetic, and biospecimen markers of PD progression. The PPMI Clinical database includes data from PD, prodromal (at risk of PD or with another form of parkinsonism), and healthy participants, who were recruited at nearly 50 international sites since 2010. The database contains a wide range of data, including demographics, medical history, clinical assessments (e.g., the Movement Disorder Society - Unified Parkinson Disease Rating Scale (MDS-UPDRS)  questionnaire, the Montreal Cognitive Assessment (MoCA) score), medication, biospecimens and proteomics, genetics, and imaging results. We used two curated datacuts obtained from PPMI upon request after approval from the PPMI Data Access Committee: \#PPMI2024 (v.20240129) comprising 3,096 patients, which was used to evaluate our approach, datacut \#PPMI2020 (v.2020420) comprising 683 patients, which was used to conduct our ablation study.

\subsubsection{ADNI}
AD is a neurodegenerative disease that leads to noticeable memory loss, cognitive decline in thinking and reasoning abilities, and significant disruptions in social or occupational functioning \cite{castellani_alzheimer_2010}. The ADNI was launched in 2003 as a public-private partnership, led by Principal Investigator Michael W. Weiner, MD. The primary goal of ADNI has been to test whether serial magnetic resonance imaging, positron emission tomography, other biological markers, and clinical and neuropsychological assessment can be combined to measure the progression of mild cognitive impairment and early AD.
More generally, ADNI data is a comprehensive and widely used collection of longitudinal clinical, imaging, genetic, and other biomarker data. It encompasses various data types, including structural, functional, and molecular brain imaging, biofluid biomarkers, cognitive assessments, genetic data, and demographic information. We used a datacut \#ADNI (ADNIMERGED\_15Oct2024) obtained from ADNI upon request after approval from the ADNI Data Access Committee.

We selected PPMI and ADNI to evaluate our approach due to their extensive use in research, with over 400 scientific publications based on its data for PPMI and 5,500 for ADNI. Of note, we intentionally selected medical datasets with a restricted access to address concerns regarding the use of LLMs trained on open-source data, not always known, and thus at risk of memorization by the LLM \cite{bordt_elephants_2024}.

\subsection{Data preparation}
To ensure the selection of a representative population of PD and AD, we included patients diagnosed with confirmed PD and AD and captured their clinical characteristics at the time of diagnosis. We then selected a subset of variables characteristic of the clinical picture of PD (and respectively, AD) and well-represented in PPMI (for PD) and ADNI (for AD) datasets, as indicated by a low missingness rate (Table VII and Table VIII in Supplementary Materials). For comparison of our approach with tabular-to-tabular models that do not handle missing values, we excluded patients with missing values. This affected the final cohort size with a loss of 19\% of patients in \#PPMI2024, 3\% patients in \#PPMI2020, and 12\% patients in \#ADNI (Table~\ref{tab: attrition chart} and Table~\ref{tab: attrition chart adni} in Appendix~\ref{apd:first}). To demonstrate the high practicality of LLMs, no preprocessing was performed on column names or values. Notably, we did not perform categorical value encoding, normalization, or outlier removal. Encoding of categorical values into their definition (e.g., "Male" or "Female" instead of 1 or 0)  would have impacted our results, as LLMs are designed to handle text and not numerical values. However, for simplicity and to ease the replication or pursue of our work, we preferred to add the categorical values and the associated mapping (e.g., 1 $\rightarrow$ "Male", 0 $\rightarrow$ "Female") in the data specifications of the prompt rather than in the preprocessing step.

\subsection{Text-to-tabular synthetic data generation}

\subsubsection{Approach}
\begin{figure}[!b]
    {\includegraphics[width=\linewidth]{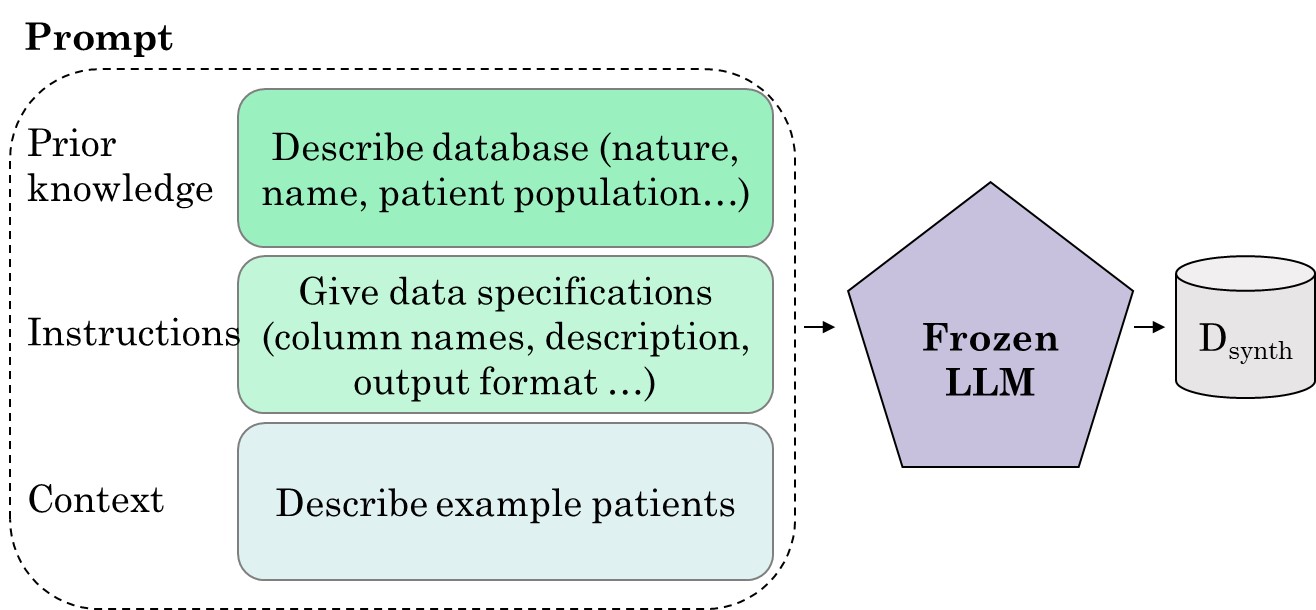}}
    {\caption{\label{fig:approach}Our text-to-tabular approach. Prior knowledge about the medical database, detailed instructions on the SDG process, and data specifications are incorporated into the prompt, along with fictitious examples of patients to provide additional context. The prompt is then fed into a frozen LLM to generate the synthetic data $D_{synth}$.}}
\end{figure}

Our approach is to prompt an LLM with a description of the desired health database in terms of nature (e.g., registry, RCT, or RWD), data specifications, sample size, and format (Figure~\ref{fig:approach}).
We used GPT-3.5 and GPT-4 through their Python API with default values of parameters and a temperature of 1. 

\subsubsection{The text - Prompt design}Our prompt used to generate synthetic data comprises several sections:
\begin{itemize}
    \item \textbf{Prior knowledge:} The prompt included prior knowledge of a medical database and a disease, that is, a comprehensive description of the medical database, detailing its name and nature, as well as the inclusion/exclusion criteria relevant to the population (e.g., disease, demographic criteria).
    \item \textbf{Instructions:} We configured the LLM to minimize verbosity by instructing it not to comment on, repeat, or answer the question. Given data specifications included detailed column names, medical definitions for each variable, the type of each variable (e.g., float, integer), and the encoding of categorical variables. The output format was specified to be in JSON, with a predetermined number of rows $n$, and no missing values.
    \item \textbf{Context:} To improve the realism of the synthetic data generated by the LLM, we leveraged its few-shot learning capabilities by including a limited number of fictitious patient examples. These examples can be derived from aggregated data found in the literature or by consulting medical experts. When using real patient records for generating examples, patient privacy must be considered. In our case, we added a single example created using the average values of the selected variables extracted from publications on \#PPMI2020 \cite{marek_parkinsons_2018} and on the ADNI dataset \cite{manning_comparison_2017, kikuchi_identification_2022}. 
\end{itemize}

The prompts designed to generate our synthetic PPMI and ADNI datasets are in the Appendix~\ref{apd:second} (see  Figure~\ref{fig:prompt_ex} for PPMI and Figure 4 in Supplementary Materials for ADNI).

\subsubsection{The tabular SDG process} To generate synthetic tabular patient data ($D_{synth}$), we simulated $m$ tables of $n$ rows using the prompt described above with $m = N/n$, $N$ being the size of the final desired database. The iterative process is mainly due to the limited size of output in terms of tokens and the laziness of LLMs which tend to truncate their response when repetitive. We noticed that $n=10$ is a reasonable choice to obtain the desired output. 
\newline 
Ordering of columns being irrelevant in tabular datasets, we randomly permuted the columns at each generation, as proposed by \cite{borisov_language_2023}.

\subsection{Evaluation framework of synthetic patient data generation} 
To obtain a comprehensive assessment, we evaluated both fidelity, i.e. statistical similarity, privacy preservation and utility, i.e.  added value of the synthetic data to a specific ML task or healthcare use case, using several widely used metrics in the SDG field or proposed by the Synthetic Data Vault in the Python library SDMetrics \cite{sdmetrics}.

\subsubsection{Evaluation framework}
We compared our approach with several baseline tabular-to-tabular SDG Models that require access to the original data during training: CTGAN \cite{xu_modeling_2019}, TVAE \cite{xu_modeling_2019}, and Gaussain copula (GC)\cite{kamthe_copula_2021}. CTGAN and TVAE were both trained for 300 epochs and using a batch size of 50. We chose a smaller batch size than the one proposed by \cite{xu_modeling_2019} because of the smaller training database. 

\begin{figure}[!b]
    \centering
    {\includegraphics[width=\linewidth]{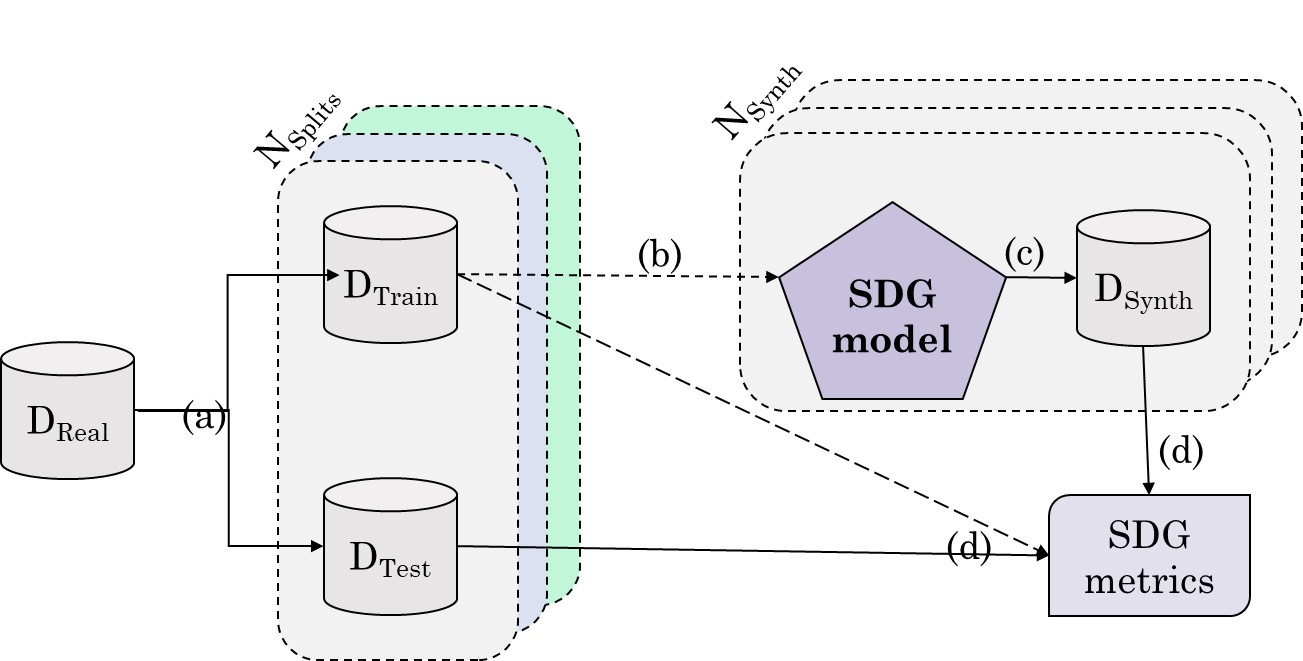}}
    {\caption{\label{fig:eval}Evaluation framework of our approach against baseline SDG Models. (a) The original dataset ($D_{\text{Real}}$) was randomly split $N_{Splits}$ times into training ($D_{\text{Train}}$, 70\%) and testing ($D_{\text{Test}}$, 30\%) sets. (b) Training of baseline models. (c)
For each split, $N_{Synth}$ different synthetic datasets ($D_{\text{Synth}}$) were generated using $D_{\text{Train}}$ (in our approach, a simple prompt is required). (d) Evaluation via computation of multiple SDG metrics averaged across data splits and synthetic data sets.}}
\end{figure}

For this comparison described in Figure~\ref{fig:eval}, we computed several metrics across five random splits of the original dataset (70\% training - $D_{Train}$, and 30\% testing - $D_{Test}$) and generated five different synthetic datasets $D_{Synth}$ for each split. For baselines, the synthetic data were generated using each training split.  We then averaged metrics across all splits and synthetic datasets, and computed the associated standard deviation. We evaluated fidelity and utility metrics between $D_{Synth}$ and $D_{Test}$ to assess the robustness of the baseline SDG Models and ensure they had not overfitted on the training data. 
We evaluated privacy metrics between $D_{Synth}$ and $D_{Train}$, considering that $D_{Train}$ was exposed to the baseline models.

\subsubsection{Fidelity metrics}To assess fidelity in terms of distribution shape, we calculated TVComplement for categorical variables (1 minus the Total Variation Distance (TVD), TVD is the distance between the frequency of each category), KSComplement for continuous variables (1 minus the Kolmogorov-Smirnov statistic, corresponding to the maximum difference between two estimated cumulative distribution functions (CDF)), and the weighted average of TVComplement and KSComplement, the Column Shape Score. Additionally, we computed two well-known distance metrics in the statistical field: the Wasserstein distance (WD) \cite{ramdas_wasserstein_2015} and then Jensen-Shannon distance (JSD) \cite{lin_divergence_1991}.

To assess fidelity in terms of correlation preservation, we calculated ContingencySimilarity (difference between contingency tables using the TVD subtracted by 1) for binary and categorical variables, CorrelationSimilarity (normalized distance between correlation matrices obtained using the Pearson correlation coefficient between each pair of variables) for continuous variables, and the associated weighted average summary metric, Column Pair Trend Score. 

Moreover, we trained a logistic regression model to distinguish the synthetic data apart from the real data, with the output metric called LogisticDetection defined as 1 minus the Area Under the Curve (AUC) score.

\subsubsection{Privacy metrics} We used two distance metrics: distance to closest record (DCR) and nearest neighbour distance ratio (NNDR) \cite{lowe_distinctive_2004}. Following \cite{zhao_ctab-gan_2021}, to provide a robust estimate of privacy preservability, we computed the fifth percentile for both metrics. Additionally was also computed the percentage of new rows generated.

Moreover, we calculated a metric named CAPCategorical measuring privacy preservation against inference attacks using the gender, the MDS-UPDRS III and MoCA score as key fields known by the "attacker" and the number of years of education as a sensitive field.

\subsubsection{Utility metric}
We trained a supervised ML model and its predictive performance on real data. More precisely, we fitted a binary classifier on $D_{Synth}$ to predict a target column and evaluated the model on $D_{test}$ using the F1-score. This metric is also usually called the Train Synthetic Test Real (TSTR) metric. Additionally, we evaluated the F1-score on the test set by adding the synthetic data to the real data during the training phase. We name this metric the Train Augmented Test Real (TATR) metric. In our case, we tried to assess whether we could predict the severity of PD and AD from baseline patient characteristics. Thus for PPMI we trained an AdaBoost model to predict the MDS-UPDRS III, a score characterizing the severity of PD motor symptoms and used to diagnose PD, which we binarized using the median value as a cut-off. Similarly, for ADNI we trained an Adaboost model to predict the mini-mental state examination (MMSE).

More detailed definitions of the computed metrics are provided in Appendix~\ref{apd:first}. Additionally, since metrics can be challenging to interpret, our evaluation included univariate and joint bivariate distribution plots from the real and synthetic datasets (see Supplementary Materials).

\subsection{Ablation study}

We conducted several ablation studies on our prompt to better understand the key components contributing to the LLM’s performance. We conducted a comparison of several experiments on the PPMI data using fidelity metrics against a reference experiment. Specifically, we examined the impact on the quality of the synthetic data of using a newer and more advanced model (by switching from GPT-3.5 to GPT-4), of adding a single example patient to the prompt, and of sampling different number of synthetic patients $n$ at each iteration in the SDG process. 

\section{Results}
\label{sec:results}

\subsection{Synthetic patient data that closely mirror the statistical properties of real data}

The computed fidelity metrics showed that, overall, the synthetic cohorts preserved the real PD and AD cohorts’ statistical properties and patterns (respectively Table~\ref{table:benchmark} and Table~\ref{table:benchmark_adni}).

The synthetic data closely resembled the real PPMI and ADNI cohorts in terms of clinical characteristics. Indeed, the comparison of distribution plots between the synthetic and real data showed close resemblances in terms of distribution shape and parameters  (Figure 5 and Figure 6 in Supplementary Materials). However, the synthetic distributions exhibited lower standard deviations and fewer outliers compared to the real data. These patterns likely reflect, respectively, the LLM's tendency to mirror prototypical PD and AD patients, and the regularizing effect of the temperature parameter.

\begin{table*}[!t]
    \caption{\label{table:benchmark}Benchmark results of SDG Models on \#PPMI2024 in terms of evaluation of fidelity, privacy preservation and utility (mean and standard deviation). For each metric, best results are reported in \textbf{bold} and second best results are \underline{underlined}. Additionally is indicated the order of magnitude of the time to generate 1000 synthetic patients (SDG time). Wassersein distance (WD), Jenssen-Shannon distance (JSD), distance to closest record (DCR), nearest neighbour distance ratio (NNDR), Train Synthetic Test Real (TSTR), Train Augmented Test Real (TATR).}
    \centering
    \begin{tabular}{p{4cm}p{2cm}p{2cm}p{2cm}p{2cm}p{2cm}}
        \toprule
        \textbf{SDG Model on \#PPMI2024 } & \textbf{$D_{train}$ (Ref.)} & \textbf{GPT-4 (Ours)} & \textbf{GC}  & \textbf{CTGAN}  & \textbf{TVAE}  \\
        \midrule
        \textbf{Fidelity} & \textbf{$D_{Test}$} & \textbf{$D_{Test}$} &  \textbf{$D_{Test}$} &  \textbf{$D_{Test}$} &  \textbf{$D_{Test}$} \\
        \midrule
        Column Shapes ($\uparrow$)&  0.956±0.017 &  0.778±0.008 &  \textbf{0.913±0.004} &  0.839±0.021 &  \underline{0.885±0.011} \\
        Column Pair Trends ($\uparrow$)  &  0.914±0.014 & 0.807±0.010 &  \textbf{0.886±0.007} & 0.851±0.009 & \underline{0.879±0.012} \\
        KSComplement ($\uparrow$)  &  0.950±0.018 & 0.767±0.011 &  \textbf{0.902±0.002} &  0.814±0.024 &  \underline{0.887±0.011} \\
        TVComplement ($\uparrow$)  &  0.978±0.014 & 0.820±0.012 &  \textbf{0.951±0.017} & \underline{0.930±0.031} &  0.877±0.031 \\
        CorrelationSimilarity ($\uparrow$)  &  0.971±0.010 & \underline{0.947±0.005} &  \textbf{0.969±0.004} &  0.917±0.007 &  0.946±0.009 \\
        ContingencySimilarity ($\uparrow$)  &  0.961±0.018 & 0.739±0.017 &  \textbf{0.916±0.019} &  \underline{0.885±0.043} &  0.801±0.044 \\
        WD ($\downarrow$)  & 0.476±0.182 & 2.274 ±0.126 &   \underline{1.023±0.077}  & 2.14±0.340 &  \textbf{1.009±0.144}\\
        JSD ($\downarrow$) &  0.183±0.007 & \textbf{0.154±0.003} &  0.195±0.004  & 0.219±0.007 & \underline{0.173±0.004} \\
        LogisticDetection ($\uparrow$) & 0.968±0.039 & 0.398±0.022 &  \textbf{0.866±0.040} &  0.405±0.118 & \underline{0.553±0.089} \\
        \midrule
        \textbf{Privacy} & \textbf{$D_{Train}$} &  \textbf{$D_{Train}$} &  \textbf{$D_{Train}$} &  \textbf{$D_{Train}$} \\
        \midrule
        NewRowSynthesis  & - & 1±0 & 1±0 & 1±0 & 1±0 \\
        DCR (5th p.) ($\uparrow$) & - & \textbf{1.556±0.038}  & 1.347±0.030 & \underline{1.391±0.042}  & 1.192±0.048  \\
        NNDR (5th p.) ($\uparrow$) & - & \underline{0.765±0.012}  & 0.727±0.016  & 0.737±0.015  & \textbf{0.768±0.016}  \\
        CategoricalCAP ($\uparrow$) & - & 0.882±0.013  & \textbf{0.911±0.010} & \underline{0.897±0.015}  & 0.857±0.017  \\
        \midrule
        \textbf{Utility} & \textbf{$D_{Test}$} & \textbf{$D_{Test}$} &  \textbf{$D_{Test}$} &  \textbf{$D_{Test}$} &  \textbf{$D_{Test}$} \\
        \midrule
        TSTR (F1 score) ($\uparrow$)& 0.978±0.112 & \textbf{0.979±0.023} & \underline{0.960±0.014} & 0.838±0.117  & 0.953±0.044 \\
         TATR (F1 score) ($\uparrow$)& - & \textbf{0.983±0.018} & \underline{0.969±0.008} & 0.883±0.093  & 0.960±0.036 \\
        \midrule
        SDG time & - & $\sim1:30hrs$ &  $\sim1s$ &  $\sim3min$ & $\sim1min$  \\
        \bottomrule
    \end{tabular}
\end{table*}

For both PPMI and ADNI cohorts, there was a noticeable shift in the mean and different distribution shapes of imaging variables: the left and right putamen variables for PPMI and the ventricles and intracranial volumes for ADNI (Figure 5 and Figure 6 in Supplementary Materials). This gap could be explained by the sensitivity of these imaging variables to the PD and AD populations and by the poor available reported statistical analysis in the literature. 

Moreover, the gender prevalence differences were notable and for PPMI the discrepancies for symptoms such as rigidity and tremor were even more pronounced when compared to the original data (Figure 5 in Supplementary Materials). These inconsistencies might be explained from the broad range of prevalence values reported in the literature, as highlighted in the meta-analysis on the overall male/female prevalence ratios in PD by \cite{zirra_gender_2022}. Furthermore, the subjective nature of the evaluation of symptoms, particularly rigidity, could contribute to these variations \cite{ferreira-sanchez_quantitative_2020}.

Finally, visual correlations between variables were well preserved (Figure 7 and Figure 8 in Supplementary Materials). 

\subsection{An approach competing with established tabular-to-tabular approaches}

Remarkably, despite not having access to the original data, our text-to-tabular approach achieved fidelity scores comparable to those of tabular-to-tabular approaches (Table~\ref{table:benchmark}). However, GC outperformed our method in terms of distribution shape and correlation.

\begin{table*}[!t]
    \centering
    {\caption{\label{table:benchmark_adni}Benchmark results of SDG Models on \#ADNI in terms of evaluation of fidelity, privacy preservation and utility (mean and standard deviation). For each metric, best results are reported in \textbf{bold} and second best results are \underline{underlined}. Additionally is indicated the order of magnitude of the time to generate 1000 synthetic patients (SDG time). Wassersein distance (WD), Jenssen-Shannon distance (JSD), distance to closest record (DCR), nearest neighbour distance ratio (NNDR), Train Synthetic Test Real (TSTR), Train Augmented Test Real (TATR).}}
    {\begin{tabular}{p{4cm}p{2cm}p{2cm}p{2cm}p{2cm}p{2cm}}
        \toprule
        \textbf{SDG Model on \#ADNI} & \textbf{$D_{train}$ (Ref.)} & \textbf{GPT-4} (Ours) & \textbf{Copula}  & \textbf{CTGAN}  & \textbf{TVAE}  \\
        \midrule
        \textbf{Fidelity} & \textbf{$D_{Test}$} & \textbf{$D_{Test}$} &  \textbf{$D_{Test}$} &  \textbf{$D_{Test}$} &  \textbf{$D_{Test}$} \\
        \midrule
        Column Shapes ($\uparrow$)&  0.922±0.006 &  0.812±0.006 &  \textbf{0.904±0.015} &  0.773±0.035 &  \underline{0.845±0.021} \\
        Column Pair Trends ($\uparrow$)  &  0.881±0.008 & 0.768±0.014 &  \textbf{0.862±0.013} & 0.806±0.012 & \underline{0.818±0.017} \\
        KSComplement ($\uparrow$)  &  0.919±0.010 & 0.785±0.010 &  \textbf{0.897±0.016} &  0.735±0.044 &  \underline{0.847±0.015} \\
        TVComplement ($\uparrow$)  &  0.938±0.020 & 0.919±0.016 &  \textbf{0.932±0.023} &  \underline{0.925±0.029} &  0.834±0.062 \\
        CorrelationSimilarity ($\uparrow$)  &  0.957±0.007 & 0.869±0.013 &  \textbf{0.954±0.011} &  0.904±0.010 &  \underline{0.942±0.006} \\
        ContingencySimilarity ($\uparrow$)  &  0.873±0.028 & \underline{0.880±0.023} &  \textbf{0.881±0.043} &  0.845±0.046 &  0.705±0.101 \\
        WD ($\downarrow$)  & 4,618±932 & 18,282±1,192 &   \textbf{6,147±472}  & 22,094±7,445 &  \underline{7,471±2,145}\\
        JSD ($\downarrow$) &  0.031±0.001 & \textbf{0.031±0.001} &  \underline{0.033±0.001}  & 0.051±0.004 & \underline{0.033±0.001} \\
        LogisticDetection ($\uparrow$) & 0.990±0.017 & 0.543±0.058 &  \textbf{0.972±0.034} &  0.315±0.146 & \underline{0.669±0.105} \\
        \midrule
        \textbf{Privacy} & \textbf{$D_{Train}$} &  \textbf{$D_{Train}$} &  \textbf{$D_{Train}$} &  \textbf{$D_{Train}$} \\
        \midrule
        NewRowSynthesis  & - & 1±0 & 1±0 & 1±0 & 1±0 \\
        DCR (5th p.) ($\uparrow$) & - & 0.967±0.030 & \underline{0.973±0.014}  & \textbf{1.064±0.048} & 0.970±0.047 \\
        NNDR (5th p.) ($\uparrow$) & -  & \underline{0.714±0.027}  & 0.674±0.023  & 0.690±0.023 & \textbf{0.718±0.029}  \\
        CategoricalCAP ($\uparrow$) & - & 0.860±0.008  & \underline{0.898±0.005} & \textbf{0.907±0.012}  & 0.883±0.012  \\
        \midrule
        \textbf{Utility} & \textbf{$D_{Test}$} & \textbf{$D_{Test}$} &  \textbf{$D_{Test}$} &  \textbf{$D_{Test}$} &  \textbf{$D_{Test}$} \\
        \midrule
        TSTR (F1 score) ($\uparrow$)& 0.933±0.091 & 0.904±0.082 & \textbf{0.947±0.066} & 0.827±0.124  & \underline{0.924±0.087} \\
        TATR (F1 score) ($\uparrow$) & - & 0.904±0.082 & \textbf{0.947±0.066} & 0.834±0.125  & \underline{0.928±0.083} \\
        \midrule
        SDG time & - & $\sim1:00hrs$ &  $\sim1s$ &  $\sim3min$ & $\sim1min$  \\
        \bottomrule
    \end{tabular}}
\end{table*}

Additionally, nearly all baseline models surpassed our approach in terms of distribution shape (except for ADNI - were GPT-4 outperformed CTGAN). Overall, GC demonstrated superior fidelity across nearly all fidelity metrics, including a LogisticDetection score, far above all others, indicating its difficulty in distinguishing real patients from synthetic ones. This may point out that GC in SDG is more efficient on smaller sample size than baseline deep learning approaches requiring more data.

Interestingly, on the PPMI data, our approach showed similar performance to TVAE and better performance than CTGAN in mimicking correlations between continuous variables. Furthermore, when applied to the ADNI dataset, our method achieved comparable performance to GC and outperformed CTGAN and TVAE in capturing correlations between discrete variables. This suggests that on certain tasks, LLMs can even compete with  traditional deep learning generative models, which is of particular interest given the lack of access to the original data.

Notably, no model exhibited significant overfitting on $D_{Train}$, as there were no substantial differences between scores computed on $D_{Train}$ and $D_{Test}$ (Table~\ref{table:benchmark2} for PPMI and Table IX in Supplementary Materials for ADNI). Our approach also excelled in preserving privacy, with the highest fifth percentile for DCR for PPMI, and the second-highest fifth percentile for NNDR after TVAE for both PPMI and ADNI (Table~\ref{table:benchmark} and Table~\ref{table:benchmark_adni}).

However, GC and CTGAN showed the highest resistance to inference attacks, with the best CAPCategorical score, though this metric is very dependent on the choice of sensitivity and key fields. Finally, on the PPMI dataset, GPT-4 achieved the highest TSTR score. Additionally, the F1-score from the binary classifier trained on our synthetic data was comparable to that of the classifier trained on the original training data, suggesting that our synthetic data could effectively replace the training data. For the ADNI dataset, while GC achieved highest TSTR score, the F1-score obtained with GPT-4 was remarkable and outperformed CTGAN.

\subsection{Ablation study}
Our ablation study showed that upgrading GPT-3.5 to GPT-4 improved the fidelity metrics (Table~\ref{tab:ablation study} in Appendix~\ref{apd:first}). Contrary to our expectation of an increase in data heterogeneity, sampling synthetic patients individually did not improve fidelity metrics.

Including an example of a typical PD patient improved the fidelity of the distribution shapes but slightly worsened the correlations. This deterioration might be attributed to the additional constraints imposed on the LLM during the generative process compared to the reference prompt.

Combining the elements that improved fidelity — upgrading to GPT-4 and adding an example — produced the best results for distribution shapes despite a slight deterioration in the fidelity of correlations.
\section{Discussion}

This study highlights the potential of LLMs in generating synthetic healthcare data without access to the original data and without requiring model fine-tuning. Our text-to-tabular approach, relying solely on a description of the desired healthcare database and a frozen LLM, showed promising results by producing realistic synthetic tabular patient data. Applied to PPMI and ADNI datasets, our method successfully generated PD and AD patients' data that preserved the characteristics of the distribution of the original data in spite of not having accessed it. Although our approach performed slightly worse in fidelity metrics compared to baseline SDG methods like GC, TVAE, and CTGAN, trained on the original data, it showed remarkable capabilities in replicating correlations between clinical variables, slightly outperforming CTGAN on some metrics. In addition, our method provides strong privacy guarantees by not using the original patient-level data. 

Despite our efforts to provide a thorough evaluation of our approach, it is worth noting that there is currently no established consensus among researchers, regulatory bodies, and the pharmaceutical industry regarding the evaluation of synthetic patient data \cite{goncalves_generation_2020}. Evaluation should be application-specific and balance the trade-off between fidelity, utility and privacy \cite{alaa_how_2022}. The level of evidence required to consider synthetic data reliable and validated for practical use should  depend on the judgment criteria and the objective of use. Indeed, the evaluation criteria for synthetic data may vary depending on whether the data is used e.g., for data augmentation or for replacing control arms in regulatory submissions.
Moreover, as highlighted in \cite{van_breugel_synthetic_2023}, users should be careful and not treat synthetic data as real, particularly when applying it to ML tasks involving underrepresented groups, as generative models are likely to struggle with accurately modeling them.

While our approach demonstrates promising results, it has certain limitations. Firstly, our evaluation was conducted on two healthcare datasets focused on well-documented diseases and reference databases. It would be valuable to assess the performance of this approach on datasets with more restrictive access, other diseases, rare diseases, and of other nature such as longitudinal data. To increase robustness of the validation, it would be valuable to extend the approach to a private benchmark of datasets or to custom data. Extending the approach to other domains beyond healthcare and evaluating it on standard benchmark tabular datasets would also strengthen the validation.

Finally, applying this approach in the appropriate setting (i.e., without access to the original patient data) may require to establish a validation framework using aggregated data from the literature or medical experts. However, it should be noted that even if the synthetic data is not fully validated, it can still be valuable for developmental purposes. Indeed, our approach has the potential to accelerate the development of ML applications in healthcare, being very flexible to the input data and able to generate within a few hours synthetic patient data at a cost that is much lower than that of traditional datasets. For instance, a ML pipeline can be implemented and tested on synthetic data before being trained and evaluated with real data once it becomes available. Furthermore, the flexibility of LLMs can be leveraged to extend the text-to-tabular approach to other modalities, such as text and graphs, thereby generating multimodal data - a focal point of current research.

\section*{Acknowledgment}

Data collection and sharing for this project was funded by the Alzheimer's Disease Neuroimaging Initiative (ADNI) (National Institutes of Health Grant U01 AG024904) and DOD ADNI (Department of Defense award number W81XWH-12-2-0012). ADNI is funded by the National Institute on Aging, the National Institute of Biomedical Imaging and
Bioengineering, and through generous contributions from the following: AbbVie, Alzheimer’s Association; Alzheimer’s Drug Discovery Foundation; Araclon Biotech; BioClinica, Inc.; Biogen; Bristol-Myers Squibb Company; CereSpir, Inc.; Cogstate; Eisai Inc.; Elan Pharmaceuticals, Inc.; Eli Lilly and Company; EuroImmun; F. Hoffmann-La Roche Ltd and its affiliated company Genentech, Inc.; Fujirebio; GE Healthcare; IXICO Ltd.; Janssen Alzheimer Immunotherapy Research \& Development, LLC.; Johnson \& Johnson Pharmaceutical Research \& Development LLC.; Lumosity; Lundbeck; Merck \& Co., Inc.; Meso Scale Diagnostics, LLC.; NeuroRx Research; Neurotrack Technologies; Novartis
Pharmaceuticals Corporation; Pfizer Inc.; Piramal Imaging; Servier; Takeda Pharmaceutical Company; and Transition Therapeutics. The Canadian Institutes of Health Research is providing funds to support ADNI clinical sites in Canada. Private sector contributions are facilitated by the Foundation for the National Institutes of Health (\url{www.fnih.org}). The grantee organization is the Northern California Institute for Research and Education, and the study is coordinated by the Alzheimer’s Therapeutic Research Institute at the University of Southern California. ADNI data are disseminated by the Laboratory for Neuro Imaging at the University of Southern California.

PPMI – a public-private partnership – is funded by the Michael J. Fox Foundation for
Parkinson’s Research and funding partners, including 4D Pharma, AbbVie Inc., AcureX Therapeutics, Allergan, Amathus Therapeutics, Aligning Science Across Parkinson’s (ASAP), Avid Radiopharmaceuticals, Bial Biotech, Biogen, BioLegend, BlueRock Therapeutics, Bristol Myers Squibb, Calico Life Sciences LLC, Celgene Corporation, DaCapo Brainscience, Denali Therapeutics, The Edmond J. Safra Foundation, Eli Lilly and Company, Gain Therapeutics, GE Healthcare, GlaxoSmithKline, Golub Capital, Handl Therapeutics, Insitro, Janssen Pharmaceuticals, Lundbeck, Merck \& Co., Inc., Meso Scale Diagnostics, LLC, Neurocrine Biosciences, Pfizer Inc., Piramal Imaging, Prevail Therapeutics, F. Hoffmann-La Roche Ltd and its affiliated company Genentech Inc., Sanofi Genzyme, Servier, Takeda Pharmaceutical Company, Teva Neuroscience, Inc., UCB, Vanqua Bio, Verily Life Sciences, Voyager Therapeutics, Inc., Yumanity Therapeutics, Inc.
4D Pharma, AbbVie Inc., AcureX Therapeutics, Allergan, Amathus Therapeutics, Aligning Science Across Parkinson’s (ASAP), Avid Radiopharmaceuticals, Bial Biotech, Biogen, BioLegend, BlueRock Therapeutics, Bristol Myers Squibb, Calico Life Sciences LLC, Celgene Corporation, DaCapo Brainscience, Denali Therapeutics, The Edmond J. Safra Foundation, Eli Lilly and Company, Gain Therapeutics, GE Healthcare, GlaxoSmithKline, Golub Capital, Handl Therapeutics, Insitro, Janssen Pharmaceuticals, Lundbeck, Merck \& Co., Inc., Meso Scale Diagnostics, LLC, Neurocrine Biosciences, Pfizer Inc., Piramal Imaging, Prevail Therapeutics, F. Hoffmann-La Roche Ltd and its affiliated company Genentech Inc., Sanofi Genzyme, Servier, Takeda Pharmaceutical Company, Teva Neuroscience, Inc., UCB, Vanqua Bio, Verily Life Sciences, Voyager Therapeutics, Inc., Yumanity Therapeutics, Inc.

\section{Supplementary Materials}
 Supplementary materials are available online at \url{https://github.com/quinten-health-os/synth-data-gen-from-text}. 
 
\bibliographystyle{IEEEtran}
\bibliography{bibliography}

\begin{thebibliography}{10}
\providecommand{\url}[1]{#1}
\csname url@samestyle\endcsname
\providecommand{\newblock}{\relax}
\providecommand{\bibinfo}[2]{#2}
\providecommand{\BIBentrySTDinterwordspacing}{\spaceskip=0pt\relax}
\providecommand{\BIBentryALTinterwordstretchfactor}{4}
\providecommand{\BIBentryALTinterwordspacing}{\spaceskip=\fontdimen2\font plus
\BIBentryALTinterwordstretchfactor\fontdimen3\font minus \fontdimen4\font\relax}
\providecommand{\BIBforeignlanguage}[2]{{%
\expandafter\ifx\csname l@#1\endcsname\relax
\typeout{** WARNING: IEEEtran.bst: No hyphenation pattern has been}%
\typeout{** loaded for the language `#1'. Using the pattern for}%
\typeout{** the default language instead.}%
\else
\language=\csname l@#1\endcsname
\fi
#2}}
\providecommand{\BIBdecl}{\relax}
\BIBdecl

\bibitem{gonzales_synthetic_2023}
\BIBentryALTinterwordspacing
A.~Gonzales, G.~Guruswamy, and S.~R. Smith, ``\BIBforeignlanguage{en}{Synthetic data in health care: {A} narrative review},'' \emph{\BIBforeignlanguage{en}{PLOS Digital Health}}, vol.~2, no.~1, p. e0000082, Jan. 2023, publisher: Public Library of Science. [Online]. Available: \url{https://journals.plos.org/digitalhealth/article?id=10.1371/journal.pdig.0000082}
\BIBentrySTDinterwordspacing

\bibitem{yale_generation_2020}
\BIBentryALTinterwordspacing
A.~Yale, S.~Dash, R.~Dutta, I.~Guyon, A.~Pavao, and K.~P. Bennett, ``Generation and evaluation of privacy preserving synthetic health data,'' \emph{Neurocomputing}, vol. 416, pp. 244--255, Nov. 2020. [Online]. Available: \url{https://www.sciencedirect.com/science/article/pii/S0925231220305117}
\BIBentrySTDinterwordspacing

\bibitem{hernandez_synthetic_2022}
\BIBentryALTinterwordspacing
M.~Hernandez, G.~Epelde, A.~Alberdi, R.~Cilla, and D.~Rankin, ``Synthetic data generation for tabular health records: {A} systematic review,'' \emph{Neurocomputing}, vol. 493, pp. 28--45, Jul. 2022. [Online]. Available: \url{https://www.sciencedirect.com/science/article/pii/S0925231222004349}
\BIBentrySTDinterwordspacing

\bibitem{eckardt_mimicking_2023}
\BIBentryALTinterwordspacing
J.-N. Eckardt, W.~Hahn, C.~Röllig, S.~Stasik, U.~Platzbecker, C.~Müller-Tidow, H.~Serve, C.~D. Baldus, C.~Schliemann, K.~Schäfer-Eckart, M.~Hanoun, M.~Kaufmann, A.~Burchert, C.~Thiede, J.~Schetelig, M.~Bornhäuser, M.~Wolfien, and J.~M. Middeke, ``Mimicking {Clinical} {Trials} with {Synthetic} {Acute} {Myeloid} {Leukemia} {Patients} {Using} {Generative} {Artificial} {Intelligence},'' \emph{Blood}, vol. 142, p. 2268, Nov. 2023. [Online]. Available: \url{https://www.sciencedirect.com/science/article/pii/S0006497123088705}
\BIBentrySTDinterwordspacing

\bibitem{perez_effectiveness_2017}
\BIBentryALTinterwordspacing
L.~Perez and J.~Wang, ``The {Effectiveness} of {Data} {Augmentation} in {Image} {Classification} using {Deep} {Learning},'' Dec. 2017, arXiv:1712.04621 [cs]. [Online]. Available: \url{http://arxiv.org/abs/1712.04621}
\BIBentrySTDinterwordspacing

\bibitem{seedat_curated_2024}
\BIBentryALTinterwordspacing
N.~Seedat, N.~Huynh, B.~van Breugel, and M.~van~der Schaar, ``Curated {LLM}: {Synergy} of {LLMs} and {Data} {Curation} for tabular augmentation in low-data regimes,'' Jun. 2024, arXiv:2312.12112 [cs]. [Online]. Available: \url{http://arxiv.org/abs/2312.12112}
\BIBentrySTDinterwordspacing

\bibitem{nakada_synthetic_2024}
\BIBentryALTinterwordspacing
R.~Nakada, Y.~Xu, L.~Li, and L.~Zhang, ``\BIBforeignlanguage{en}{Synthetic {Oversampling}: {Theory} and {A} {Practical} {Approach} {Using} {LLMs} to {Address} {Data} {Imbalance}},'' Jun. 2024, arXiv:2406.03628 [cs, stat]. [Online]. Available: \url{http://arxiv.org/abs/2406.03628}
\BIBentrySTDinterwordspacing

\bibitem{pauley_t1dctegui_2023}
\BIBentryALTinterwordspacing
M.~Pauley, N.~Henscheid, S.~E. David, S.~R. Karpen, K.~Romero, J.~T. Podichetty, and t.~T. .~D. Consortium~(T1DC), ``\BIBforeignlanguage{en}{{T1dCteGui}: {A} {User}-{Friendly} {Clinical} {Trial} {Enrichment} {Tool} to {Optimize} {T1D} {Prevention} {Studies} by {Leveraging} {AI}/{ML} {Based} {Synthetic} {Patient} {Population}},'' \emph{\BIBforeignlanguage{en}{Clinical Pharmacology \& Therapeutics}}, vol. 114, no.~3, pp. 704--711, 2023, \_eprint: https://onlinelibrary.wiley.com/doi/pdf/10.1002/cpt.2976. [Online]. Available: \url{https://onlinelibrary.wiley.com/doi/abs/10.1002/cpt.2976}
\BIBentrySTDinterwordspacing

\bibitem{lhostis_knowledge-based_2023}
\BIBentryALTinterwordspacing
A.~L’Hostis, J.-L. Palgen, A.~Perrillat-Mercerot, E.~Peyronnet, E.~Jacob, J.~Bosley, M.~Duruisseaux, R.~Toueg, L.~Lefèvre, R.~Kahoul, N.~Ceres, and C.~Monteiro, ``\BIBforeignlanguage{en}{Knowledge-based mechanistic modeling accurately predicts disease progression with gefitinib in {EGFR}-mutant lung adenocarcinoma},'' \emph{\BIBforeignlanguage{en}{npj Systems Biology and Applications}}, vol.~9, no.~1, p.~37, Jul. 2023. [Online]. Available: \url{https://www.nature.com/articles/s41540-023-00292-7}
\BIBentrySTDinterwordspacing

\bibitem{alam_digital_2024}
\BIBentryALTinterwordspacing
N.~Alam, J.~Basilico, D.~Bertolini, S.~C. Chetty, H.~D'Angelo, R.~Douglas, C.~K. Fisher, F.~Fuller, M.~Gomes, R.~Gupta, A.~Lang, A.~Loukianov, R.~Mak-McCully, C.~Murray, H.~Pham, S.~Qiao, E.~Ryapolova-Webb, A.~Smith, D.~Theoharatos, A.~Tolwani, E.~W. Tramel, A.~Vidovszky, J.~Viduya, and J.~R. Walsh, ``\BIBforeignlanguage{en}{Digital {Twin} {Generators} for {Disease} {Modeling}},'' May 2024, arXiv:2405.01488 [cs, stat]. [Online]. Available: \url{http://arxiv.org/abs/2405.01488}
\BIBentrySTDinterwordspacing

\bibitem{borisov_language_2023}
\BIBentryALTinterwordspacing
V.~Borisov, K.~Seßler, T.~Leemann, M.~Pawelczyk, and G.~Kasneci, ``\BIBforeignlanguage{en}{Language {Models} are {Realistic} {Tabular} {Data} {Generators}},'' Apr. 2023, arXiv:2210.06280 [cs]. [Online]. Available: \url{http://arxiv.org/abs/2210.06280}
\BIBentrySTDinterwordspacing

\bibitem{borisov_deep_2024}
\BIBentryALTinterwordspacing
V.~Borisov, T.~Leemann, K.~Seßler, J.~Haug, M.~Pawelczyk, and G.~Kasneci, ``\BIBforeignlanguage{en}{Deep {Neural} {Networks} and {Tabular} {Data}: {A} {Survey}},'' \emph{\BIBforeignlanguage{en}{IEEE Transactions on Neural Networks and Learning Systems}}, vol.~35, no.~6, pp. 7499--7519, Jun. 2024, arXiv:2110.01889 [cs]. [Online]. Available: \url{http://arxiv.org/abs/2110.01889}
\BIBentrySTDinterwordspacing

\bibitem{young_using_2009}
J.~Young, P.~Graham, and R.~Penny, ``Using {Bayesian} {Networks} to {Create} {Synthetic} {Data},'' \emph{Journal of Official Statistics}, vol.~25, pp. 549--567, Dec. 2009.

\bibitem{zhang_privbayes_nodate}
\BIBentryALTinterwordspacing
J.~Zhang, G.~Cormode, C.~M. Procopiuc, D.~Srivastava, and X.~Xiao, ``\BIBforeignlanguage{en}{{PrivBayes}: {Private} {Data} {Release} via {Bayesian} {Networks}},'' \emph{\BIBforeignlanguage{en}{Proceedings of the 2014 ACM SIGMOD International Conference on Management of Data}}, p. 1423–1434, 2014. [Online]. Available: \url{https://doi.org/10.1145/2588555.2588573}
\BIBentrySTDinterwordspacing

\bibitem{patki_synthetic_2016}
\BIBentryALTinterwordspacing
N.~Patki, R.~Wedge, and K.~Veeramachaneni, ``\BIBforeignlanguage{en}{The {Synthetic} {Data} {Vault}},'' in \emph{\BIBforeignlanguage{en}{2016 {IEEE} {International} {Conference} on {Data} {Science} and {Advanced} {Analytics} ({DSAA})}}.\hskip 1em plus 0.5em minus 0.4em\relax Montreal, QC, Canada: IEEE, Oct. 2016, pp. 399--410. [Online]. Available: \url{http://ieeexplore.ieee.org/document/7796926/}
\BIBentrySTDinterwordspacing

\bibitem{kamthe_copula_2021}
\BIBentryALTinterwordspacing
S.~Kamthe, S.~Assefa, and M.~Deisenroth, ``Copula {Flows} for {Synthetic} {Data} {Generation},'' Jan. 2021, arXiv:2101.00598 [cs, stat]. [Online]. Available: \url{http://arxiv.org/abs/2101.00598}
\BIBentrySTDinterwordspacing

\bibitem{goodfellow_generative_2014}
\BIBentryALTinterwordspacing
I.~J. Goodfellow, J.~Pouget-Abadie, M.~Mirza, B.~Xu, D.~Warde-Farley, S.~Ozair, A.~Courville, and Y.~Bengio, ``Generative {Adversarial} {Networks},'' Jun. 2014, arXiv:1406.2661 [cs, stat]. [Online]. Available: \url{http://arxiv.org/abs/1406.2661}
\BIBentrySTDinterwordspacing

\bibitem{xu_modeling_2019}
\BIBentryALTinterwordspacing
L.~Xu, M.~Skoularidou, A.~Cuesta-Infante, and K.~Veeramachaneni, ``\BIBforeignlanguage{en}{Modeling {Tabular} data using {Conditional} {GAN}},'' Oct. 2019, arXiv:1907.00503 [cs, stat]. [Online]. Available: \url{http://arxiv.org/abs/1907.00503}
\BIBentrySTDinterwordspacing

\bibitem{zhao_ctab-gan_2021}
\BIBentryALTinterwordspacing
Z.~Zhao, A.~Kunar, H.~Van~der Scheer, R.~Birke, and L.~Y. Chen, ``{CTAB}-{GAN}: {Effective} {Table} {Data} {Synthesizing},'' May 2021, arXiv:2102.08369 [cs]. [Online]. Available: \url{http://arxiv.org/abs/2102.08369}
\BIBentrySTDinterwordspacing

\bibitem{xie_differentially_2018}
\BIBentryALTinterwordspacing
L.~Xie, K.~Lin, S.~Wang, F.~Wang, and J.~Zhou, ``Differentially {Private} {Generative} {Adversarial} {Network},'' Feb. 2018, arXiv:1802.06739 [cs, stat]. [Online]. Available: \url{http://arxiv.org/abs/1802.06739}
\BIBentrySTDinterwordspacing

\bibitem{jordon_pate-gan_2018}
\BIBentryALTinterwordspacing
J.~Jordon, J.~Yoon, and M.~v.~d. Schaar, ``\BIBforeignlanguage{en}{{PATE}-{GAN}: {Generating} {Synthetic} {Data} with {Differential} {Privacy} {Guarantees}},'' in \emph{\BIBforeignlanguage{en}{International Conference on Learning}}, Sep. 2018. [Online]. Available: \url{https://openreview.net/forum?id=S1zk9iRqF7}
\BIBentrySTDinterwordspacing

\bibitem{zhao_ctab-gan_2022}
\BIBentryALTinterwordspacing
Z.~Zhao, A.~Kunar, R.~Birke, and L.~Y. Chen, ``{CTAB}-{GAN}+: {Enhancing} {Tabular} {Data} {Synthesis},'' \emph{Frontiers in Big Data}, vol.~6, 2022, publisher: arXiv Version Number: 1. [Online]. Available: \url{https://arxiv.org/abs/2204.00401}
\BIBentrySTDinterwordspacing

\bibitem{austin_structured_2021}
\BIBentryALTinterwordspacing
J.~Austin, D.~D. Johnson, J.~Ho, D.~Tarlow, and R.~van~den Berg, ``Structured {Denoising} {Diffusion} {Models} in {Discrete} {State}-{Spaces},'' in \emph{Advances in {Neural} {Information} {Processing} {Systems}}, vol.~34.\hskip 1em plus 0.5em minus 0.4em\relax Curran Associates, Inc., 2021, pp. 17\,981--17\,993. [Online]. Available: \url{https://proceedings.neurips.cc/paper/2021/hash/958c530554f78bcd8e97125b70e6973d-Abstract.html}
\BIBentrySTDinterwordspacing

\bibitem{kotelnikov_tabddpm_2022}
\BIBentryALTinterwordspacing
A.~Kotelnikov, D.~Baranchuk, I.~Rubachev, and A.~Babenko, ``{TabDDPM}: {Modelling} {Tabular} {Data} with {Diffusion} {Models},'' Sep. 2022, arXiv:2209.15421 [cs]. [Online]. Available: \url{http://arxiv.org/abs/2209.15421}
\BIBentrySTDinterwordspacing

\bibitem{jolicoeur-martineau_generating_2024}
\BIBentryALTinterwordspacing
A.~Jolicoeur-Martineau, K.~Fatras, and T.~Kachman, ``\BIBforeignlanguage{en}{Generating and {Imputing} {Tabular} {Data} via {Diffusion} and {Flow}-based {Gradient}-{Boosted} {Trees}},'' in \emph{\BIBforeignlanguage{en}{Proceedings of {The} 27th {International} {Conference} on {Artificial} {Intelligence} and {Statistics}}}.\hskip 1em plus 0.5em minus 0.4em\relax PMLR, Apr. 2024, pp. 1288--1296, iSSN: 2640-3498. [Online]. Available: \url{https://proceedings.mlr.press/v238/jolicoeur-martineau24a.html}
\BIBentrySTDinterwordspacing

\bibitem{solatorio_realtabformer_2023}
\BIBentryALTinterwordspacing
A.~V. Solatorio and O.~Dupriez, ``{REaLTabFormer}: {Generating} {Realistic} {Relational} and {Tabular} {Data} using {Transformers},'' 2023, publisher: arXiv Version Number: 1. [Online]. Available: \url{https://arxiv.org/abs/2302.02041}
\BIBentrySTDinterwordspacing

\bibitem{zhao_tabula_2023}
\BIBentryALTinterwordspacing
Z.~Zhao, R.~Birke, and L.~Chen, ``{TabuLa}: {Harnessing} {Language} {Models} for {Tabular} {Data} {Synthesis},'' 2023, publisher: arXiv Version Number: 1. [Online]. Available: \url{https://arxiv.org/abs/2310.12746}
\BIBentrySTDinterwordspacing

\bibitem{tran_differentially_2024}
\BIBentryALTinterwordspacing
T.~V. Tran and L.~Xiong, ``\BIBforeignlanguage{en}{Differentially {Private} {Tabular} {Data} {Synthesis} using {Large} {Language} {Models}},'' Jun. 2024, arXiv:2406.01457 [cs]. [Online]. Available: \url{http://arxiv.org/abs/2406.01457}
\BIBentrySTDinterwordspacing

\bibitem{ling_mallm-gan_2024}
\BIBentryALTinterwordspacing
Y.~Ling, X.~Jiang, and Y.~Kim, ``\BIBforeignlanguage{en}{{MALLM}-{GAN}: {Multi}-{Agent} {Large} {Language} {Model} as {Generative} {Adversarial} {Network} for {Synthesizing} {Tabular} {Data}},'' Jun. 2024, arXiv:2406.10521 [cs]. [Online]. Available: \url{http://arxiv.org/abs/2406.10521}
\BIBentrySTDinterwordspacing

\bibitem{xu_are_2024}
\BIBentryALTinterwordspacing
S.~Xu, C.-T. Lee, M.~Sharma, R.~B. Yousuf, N.~Muralidhar, and N.~Ramakrishnan, ``\BIBforeignlanguage{en}{Are {LLMs} {Naturally} {Good} at {Synthetic} {Tabular} {Data} {Generation}?}'' Jun. 2024, arXiv:2406.14541 [cs]. [Online]. Available: \url{http://arxiv.org/abs/2406.14541}
\BIBentrySTDinterwordspacing

\bibitem{cook_creating_2024}
\BIBentryALTinterwordspacing
D.~A. Cook, ``\BIBforeignlanguage{EN}{Creating virtual patients using large language models: scalable, global, and low cost},'' \emph{\BIBforeignlanguage{EN}{Medical Teacher}}, Jul. 2024, publisher: Taylor \& Francis. [Online]. Available: \url{https://www.tandfonline.com/doi/abs/10.1080/0142159X.2024.2376879}
\BIBentrySTDinterwordspacing

\bibitem{politis_parkinsons_2010}
\BIBentryALTinterwordspacing
M.~Politis, K.~Wu, S.~Molloy, P.~G.~Bain, K.~R. Chaudhuri, and P.~Piccini, ``\BIBforeignlanguage{en}{Parkinson's disease symptoms: {The} patient's perspective},'' \emph{\BIBforeignlanguage{en}{Movement Disorders}}, vol.~25, no.~11, pp. 1646--1651, 2010, \_eprint: https://onlinelibrary.wiley.com/doi/pdf/10.1002/mds.23135. [Online]. Available: \url{https://onlinelibrary.wiley.com/doi/abs/10.1002/mds.23135}
\BIBentrySTDinterwordspacing

\bibitem{castellani_alzheimer_2010}
\BIBentryALTinterwordspacing
R.~J. Castellani, R.~K. Rolston, and M.~A. Smith, ``\BIBforeignlanguage{en}{Alzheimer {Disease}},'' \emph{\BIBforeignlanguage{en}{Disease-a-month : DM}}, vol.~56, no.~9, p. 484, Sep. 2010. [Online]. Available: \url{https://pmc.ncbi.nlm.nih.gov/articles/PMC2941917/}
\BIBentrySTDinterwordspacing

\bibitem{bordt_elephants_2024}
\BIBentryALTinterwordspacing
S.~Bordt, H.~Nori, and R.~Caruana, ``Elephants {Never} {Forget}: {Testing} {Language} {Models} for {Memorization} of {Tabular} {Data},'' Mar. 2024, arXiv:2403.06644 [cs]. [Online]. Available: \url{http://arxiv.org/abs/2403.06644}
\BIBentrySTDinterwordspacing

\bibitem{marek_parkinsons_2018}
\BIBentryALTinterwordspacing
K.~Marek, S.~Chowdhury, A.~Siderowf, S.~Lasch, C.~S. Coffey, C.~Caspell-Garcia, T.~Simuni, D.~Jennings, C.~M. Tanner, J.~Q. Trojanowski, L.~M. Shaw, J.~Seibyl, N.~Schuff, A.~Singleton, K.~Kieburtz, A.~W. Toga, B.~Mollenhauer, D.~Galasko, L.~M. Chahine, D.~Weintraub, T.~Foroud, D.~Tosun-Turgut, K.~Poston, V.~Arnedo, M.~Frasier, T.~Sherer, and t.~P. P.~M. Initiative, ``\BIBforeignlanguage{en}{The {Parkinson}'s progression markers initiative ({PPMI}) – establishing a {PD} biomarker cohort},'' \emph{\BIBforeignlanguage{en}{Annals of Clinical and Translational Neurology}}, vol.~5, no.~12, pp. 1460--1477, 2018, \_eprint: https://onlinelibrary.wiley.com/doi/pdf/10.1002/acn3.644. [Online]. Available: \url{https://onlinelibrary.wiley.com/doi/abs/10.1002/acn3.644}
\BIBentrySTDinterwordspacing

\bibitem{manning_comparison_2017}
\BIBentryALTinterwordspacing
E.~N. Manning, K.~K. Leung, J.~M. Nicholas, I.~B. Malone, M.~J. Cardoso, J.~M. Schott, N.~C. Fox, and J.~Barnes, ``\BIBforeignlanguage{en}{A {Comparison} of {Accelerated} and {Non}-accelerated {MRI} {Scans} for {Brain} {Volume} and {Boundary} {Shift} {Integral} {Measures} of {Volume} {Change}: {Evidence} from the {ADNI} {Dataset}},'' \emph{\BIBforeignlanguage{en}{Neuroinformatics}}, vol.~15, no.~2, pp. 215--226, Apr. 2017, company: Springer Distributor: Springer Institution: Springer Label: Springer Number: 2 Publisher: Springer US. [Online]. Available: \url{https://link.springer.com/article/10.1007/s12021-017-9326-0}
\BIBentrySTDinterwordspacing

\bibitem{kikuchi_identification_2022}
\BIBentryALTinterwordspacing
M.~Kikuchi, K.~Kobayashi, S.~Itoh, K.~Kasuga, A.~Miyashita, T.~Ikeuchi, E.~Yumoto, Y.~Kosaka, Y.~Fushimi, T.~Takeda, S.~Manabe, S.~Hattori, A.~D.~N. Initiative, A.~Nakaya, K.~Kamijo, and Y.~Matsumura, ``\BIBforeignlanguage{en}{Identification of mild cognitive impairment subtypes predicting conversion to {Alzheimer}’s disease using multimodal data},'' \emph{\BIBforeignlanguage{en}{Computational and Structural Biotechnology Journal}}, vol.~20, p. 5296, Aug. 2022. [Online]. Available: \url{https://pmc.ncbi.nlm.nih.gov/articles/PMC9513733/}
\BIBentrySTDinterwordspacing

\bibitem{sdmetrics}
\BIBentryALTinterwordspacing
\emph{Synthetic Data Metrics}, DataCebo, Inc., 5 2024, version 0.14.1. [Online]. Available: \url{https://docs.sdv.dev/sdmetrics/}
\BIBentrySTDinterwordspacing

\bibitem{ramdas_wasserstein_2015}
\BIBentryALTinterwordspacing
A.~Ramdas, N.~Garcia, and M.~Cuturi, ``On {Wasserstein} {Two} {Sample} {Testing} and {Related} {Families} of {Nonparametric} {Tests},'' Oct. 2015, arXiv:1509.02237 [math, stat]. [Online]. Available: \url{http://arxiv.org/abs/1509.02237}
\BIBentrySTDinterwordspacing

\bibitem{lin_divergence_1991}
\BIBentryALTinterwordspacing
J.~Lin, ``Divergence measures based on the {Shannon} entropy,'' \emph{IEEE Transactions on Information Theory}, vol.~37, no.~1, pp. 145--151, Jan. 1991, conference Name: IEEE Transactions on Information Theory. [Online]. Available: \url{https://ieeexplore.ieee.org/document/61115}
\BIBentrySTDinterwordspacing

\bibitem{lowe_distinctive_2004}
\BIBentryALTinterwordspacing
D.~G. Lowe, ``\BIBforeignlanguage{en}{Distinctive {Image} {Features} from {Scale}-{Invariant} {Keypoints}},'' \emph{\BIBforeignlanguage{en}{International Journal of Computer Vision}}, vol.~60, no.~2, pp. 91--110, Nov. 2004. [Online]. Available: \url{https://doi.org/10.1023/B:VISI.0000029664.99615.94}
\BIBentrySTDinterwordspacing

\bibitem{zirra_gender_2022}
\BIBentryALTinterwordspacing
A.~Zirra, S.~C. Rao, J.~Bestwick, R.~Rajalingam, C.~Marras, C.~Blauwendraat, I.~F. Mata, and A.~J. Noyce, ``Gender {Differences} in the {Prevalence} of {Parkinson}'s {Disease},'' \emph{Movement Disorders Clinical Practice}, vol.~10, no.~1, pp. 86--93, Nov. 2022. [Online]. Available: \url{https://www.ncbi.nlm.nih.gov/pmc/articles/PMC9847309/}
\BIBentrySTDinterwordspacing

\bibitem{ferreira-sanchez_quantitative_2020}
\BIBentryALTinterwordspacing
M.~d.~R. Ferreira-Sánchez, M.~Moreno-Verdú, and R.~Cano-de-la Cuerda, ``Quantitative {Measurement} of {Rigidity} in {Parkinson}’s {Disease}: {A} {Systematic} {Review},'' \emph{Sensors (Basel, Switzerland)}, vol.~20, no.~3, p. 880, Feb. 2020. [Online]. Available: \url{https://www.ncbi.nlm.nih.gov/pmc/articles/PMC7038663/}
\BIBentrySTDinterwordspacing

\bibitem{goncalves_generation_2020}
\BIBentryALTinterwordspacing
A.~Goncalves, P.~Ray, B.~Soper, J.~Stevens, L.~Coyle, and A.~P. Sales, ``\BIBforeignlanguage{en}{Generation and evaluation of synthetic patient data},'' \emph{\BIBforeignlanguage{en}{BMC Medical Research Methodology}}, vol.~20, no.~1, p. 108, Dec. 2020. [Online]. Available: \url{https://bmcmedresmethodol.biomedcentral.com/articles/10.1186/s12874-020-00977-1}
\BIBentrySTDinterwordspacing

\bibitem{alaa_how_2022}
\BIBentryALTinterwordspacing
A.~M. Alaa, B.~van Breugel, E.~Saveliev, and M.~van~der Schaar, ``How {Faithful} is your {Synthetic} {Data}? {Sample}-level {Metrics} for {Evaluating} and {Auditing} {Generative} {Models},'' Jul. 2022, arXiv:2102.08921 [cs, stat]. [Online]. Available: \url{http://arxiv.org/abs/2102.08921}
\BIBentrySTDinterwordspacing

\bibitem{van_breugel_synthetic_2023}
\BIBentryALTinterwordspacing
B.~van Breugel, Z.~Qian, and M.~van~der Schaar, ``\BIBforeignlanguage{en}{Synthetic data, real errors: how (not) to publish and use synthetic data},'' Jul. 2023, arXiv:2305.09235 [cs]. [Online]. Available: \url{http://arxiv.org/abs/2305.09235}
\BIBentrySTDinterwordspacing

\end{thebibliography}

\appendices

\section{}\label{apd:first}

 \subsection{Data}
 \subsubsection{PPMI}
Data used in the preparation of this article were obtained on February, 6th, 2024 from the
Parkinson’s Progression Markers Initiative (PPMI) database (www.ppmi-info.org/access-dataspecimens/download-data), RRID:SCR 006431. For up-to-date information on the study, visit www.ppmi-info.org.

\begin{table}[htbp]
    \centering
    {\caption{\label{tab: attrition chart}Attrition chart of PPMI study populations}}
    {
    \begin{tabular}{p{2cm}p{2cm}p{2cm}}
        \toprule
        & \textbf{\#PPMI2024} & \textbf{\#PPMI2020}\\
        \midrule
        Total & 3,096 & 683\\
        \addlinespace
        PD & 1,295 & 423\\
        \addlinespace
        Preprocessing & 1,052 & 410 \\
        \bottomrule
    \end{tabular}}
\end{table}

\subsubsection{ADNI}

Data used in preparation of this article were obtained from the Alzheimer’s Disease Neuroimaging Initiative (ADNI) database (adni.loni.usc.edu). As such, the investigators within the ADNI contributed to the design and implementation of ADNI and/or provided data
but did not participate in analysis or writing of this report. A complete listing of ADNI investigators can be found at: \url{https://adni.loni.usc.edu/wp-content/uploads/how_to_apply/ADNI_Acknowledgement_List.pdf}

\begin{table}[htbp]
    \centering
    {\caption{\label{tab: attrition chart adni}Attrition chart of ADNI study population}}
    {
    \begin{tabular}{p{2cm}p{2cm}}
        \toprule
        & \textbf{\#ADNI} \\
        \midrule
        Total & 2,430\\
        \addlinespace
        AD & 411\\
        \addlinespace
        Preprocessing & 362 \\
        \bottomrule
    \end{tabular}}
\end{table}

\subsection{Data generation}
We used GPT-3.5 (gpt-3.5-turbo-0125) and GPT-4 (gpt-4-turbo-2024-04-09) through their Python API with default value parameters and a temperature of 1.

\subsection{Evaluation}
\subsubsection{Fidelity} 

\paragraph{Distribution plots}
\begin{itemize}
    \item \textbf{Univariate distributions plots:} We compared distributions one by one of synthetic data against real data to examine the distribution shape and identify potential issues in the patterns learned by the SDG Model.
    \item \textbf{Joint bivariate distribution plots:} We compared scatter plots of pairs of continuous distributions to examine the relations that were maintained between the original and synthetic data.
\end{itemize}

\paragraph{Univariate similarity  metrics}
\begin{itemize}
    \item \textbf{KSComplement:} similarity metric using the Kolmogoriv-Smirnov (KS) statistic corresponding to the maximum difference between two cumulative distribution functions (CDF). KSComplement is computed as inverting the KS statistic (i.e. 1 – KS statistic).
    \item \textbf{TVComplement:} similarity metric using the total variation distance (TVD) between two discrete or categorical distributions. TVD is the distance between the frequency of each category values within the two distributions.
\end{itemize}

Both KSComplement and TVComplement metric values are bounded between 0 and 1. The value of 0 corresponds to real and synthetic data that are as different as they can be while a value of 1 corresponds to a perfect overlap between the real and synthetic distributions.

\begin{itemize}
    \item \textbf{Column Shape Score:} summary metric provided by the SDV library, is computed as the weighted average of the KSComplement and TVComplement scores \cite{sdmetrics}.
    \item \textbf{Jensen-Shannon distance (JSD) \cite{lin_divergence_1991}:} distance computed as the square root of the Jensen-Shannon divergence, corresponding to a symmetrized and smoothed version of the Kullback–Leibler divergence; a measure of difference between two probability distributions. The distance ranges from 0 to 1 with 0 indicating that the two distributions are identical.
    \item \textbf{Wasserstein distance (WD) \cite{ramdas_wasserstein_2015}:} also known as the earthmover distance, is a similarity metric that quantifies the amount of “work” (i.e. how much mass must be moved around, and how far) needed to transform one distribution into another. In 2D, it corresponds to the area between cumulative distribution functions. Values ranges between 0 and infinity, 0 corresponding to perfect overlap between the two distributions.
    \item \textbf{LogisticDetection}: metric defined as 1 minus the Area Under the Curve (AUC) of a logistic regression classifier model trained to detect synthetic data apart from real data. The classifier's performance was evaluated using cross-validation. The classifier used was scikit-learn's LogisticRegression - the default classifier from the sdvmetrics library, with default parameters.
\end{itemize}

\paragraph{Multivariate similarity metrics}
\begin{itemize}
    \item \textbf{ContingencySimilarity:} similarity metric between a pair of categorical or binary variables between the synthetic and real data. Corresponds to the difference between contingency tables using the TVD subtracted by 1. 

    \item \textbf{CorrelationSimilarity:} similarity metric between a pair of continuous variables between the synthetic and real data. Computes a normalized distance between correlation matrices obtained using the Pearson correlation coefficient between each pair of variables.
\end{itemize}

Both ContingencySimilarity and CorrelationSimilarity metric values range between 0 and 1. The value of 1 corresponding to a perfect overlap between contingency or correlation matrices of the original and synthetic data.

\begin{itemize}
    \item \textbf{Column Pair Trend Score:} summary metric provided by the SDV library that is computed as the average of the ContingencySimilarity or CorrelationSimilarity scores of pair of variables. For pairs of continuous and discrete variables, the continuous variables are discretized into categories and then the ContingencySimilarity is computed. 
\end{itemize}

\subsubsection{Privacy preservation}
\begin{itemize}
    \item \textbf{Distance to closest record (DCR):} Distance to closest record measures the closest distance between a point in the synthetic data and the real data. The Euclidean distance is computed for each point of the synthetic data and then averaged. As suggested by \cite{zhao_ctab-gan_2021}, the 5th percentile of the metric was computed to provide a robust estimate of the privacy risk. 

    \item \textbf{Nearest neighbor distance ratio (NNDR) \cite{lowe_distinctive_2004}:} NNDR is a metric used to assess the proximity of a data point to outliers in the real data. It is defined as the ratio of the Euclidean distance to the nearest neighbor to the distance to the second nearest neighbor. An NNDR value close to 1 indicates that the point is likely situated within a cluster of the real data, whereas a value close to 0 suggests proximity to an outlier. By calculating NNDR for each synthetic record within the real data, we can identify whether the synthetic points are located in sparsely populated regions or within densely populated regions of the real data space. As well as for the DCR, the 5th percentile was provided for this metric.

    \item \textbf{CAPCategorical:} metric that evaluates the risk of revealing confidential information via an inference attack. More precisely, it evaluates the probability of an attacker correctly inferring sensitive information by combining known values from the real data with synthetic data using the correct attribution probability (CAP) algorithm. A value of 1 correspond to the original data being 100\% safe from an attack while a value of 0 correspond to the attacker being able to guess every sensitive value given some information from the real data. In our case, the selected key fields were sex, updrs3 score, moca and the sensitive field EDUCYRS.
    \item \textbf{NewRowSynthesis:} percentage of new rows in the synthetic data compared to the original data.
\end{itemize}

\subsubsection{Utility}
\begin{itemize}
    \item \textbf{BinaryAdaboostClassifier}
    To evaluate the added value of using synthetic data for a ML binary prediction task, an AdaBoost classifier is fitted to the training data. The evaluation process involves training the ML algorithm with synthetic data. The efficacy of the ML model is then assessed by making predictions on the test data. The final assessment metric is the F1 test score in this case also called the Train Synthetic Test Real (TSTR) score.

    \item \textbf{BinaryAdaboostClassifier - augmented}
    To evaluate the added value of augmenting the real data with synthetic data for a ML binary prediction task, an AdaBoost classifier is fitted to both synthetic and a part of the real data. The efficacy of the ML model is then assessed by making predictions on the test data, a hold-out part of the real data. The final assessment metric is the F1 test score and should be compared to the F1-score obtained with the same model trained solely on the real data. The final assessment metric is the F1 test score which we call the Train Augmented Test Real (TATR) score.

\end{itemize}

\subsection{Ablation results}
\begin{sidewaystable}
    \centering
    {\caption{\label{tab:ablation study}Ablation study}}
    {\begin{tabular}{lp{1.5cm}p{0.5cm}p{1cm}p{1cm}p{1cm}p{1cm}p{1cm}p{1cm}p{1cm}}
    \toprule
    Experience & Model & \textit{n} & Example row  & Column shape  & Column pair trend  & KS Compl. & TV Compl. & Corr. Sim. & Cont. Sim. \\
    \midrule
    Reference & GPT-3.5 & 10 &  & 0.64 & \textbf{0.84} & 0.58 & 0.88 & 0.94 & 0.79 \\
    Updated LLM & GPT-4 & 10 & & 0.71 & 0.83 & \textbf{0.68} & 0.80 & \textbf{0.95} & 0.75 \\
    Sampling 1 by 1 & GPT-3.5 & 1 & & 0.55 & 0.77 & 0.48 & 0.83 & 0.94 & 0.70 \\
    Example row added& GPT-3.5 & 10 & \checkmark & \textbf{0.70} & 0.82 & 0.64 & \textbf{0.94} & 0.91 & \textbf{0.90} \\
    Best parameters & GPT-4 & 10 &\checkmark  & 0.76 & 0.82 & 0.74 & 0.81 & 0.94 & 0.75 \\
    \bottomrule
    \end{tabular}}
\end{sidewaystable}
\section{}\label{apd:second}

\begin{figure*}[!ht]
    \centering
    {\caption{\label{fig:prompt_ex}Prompt used to generate a synthetic Parkinson's disease population from PPMI.}}
    {\includegraphics[width=\linewidth]{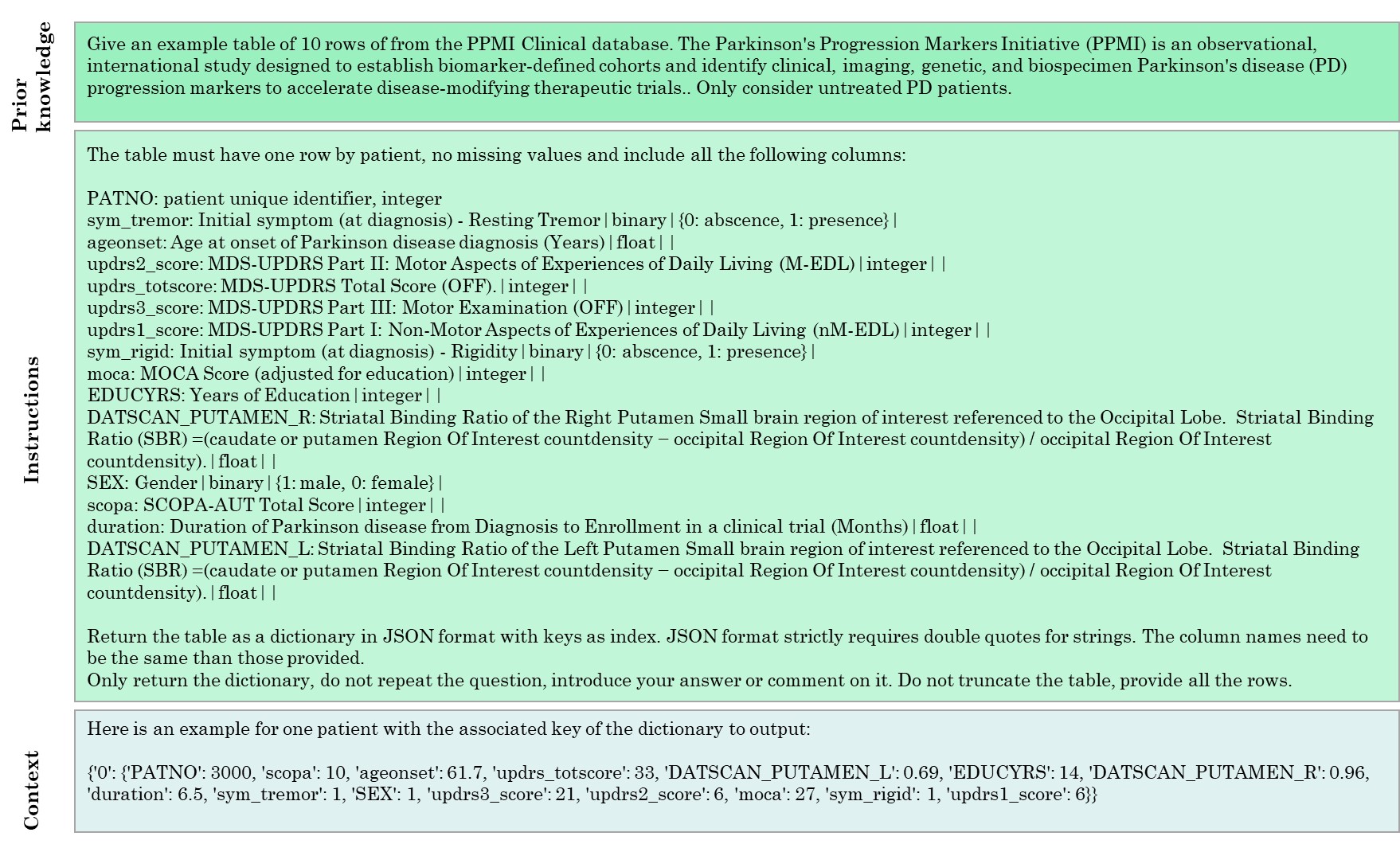}}
\end{figure*}

\begin{table*}[!t]
    \centering
    {\caption{\label{table:benchmark2}Benchmark results of SDG Models on \#PPMI2024 in terms of evaluation of fidelity, privacy preservation and utility (mean and standard deviation). For each metric, the best results are reported in bold.}}
    \begin{tabular}{p{3.2cm}p{1.5cm}p{1.5cm}p{1.5cm}p{1.5cm}p{1.5cm}p{1.5cm}p{1.5cm}p{1.5cm}}
        \toprule
        \textbf{} & \textbf{Ours} & \textbf{} & \textbf{Baselines} & \textbf{} & \textbf{} & \textbf{} & \textbf{} & \textbf{} \\
        \midrule
        \textbf{Model} & \textbf{GPT-4} & \textbf{} & \textbf{GC} & \textbf{} & \textbf{CTGAN} & \textbf{} & \textbf{TVAE} & \textbf{} \\
        \textbf{Metric} & \textbf{$D_{Train}$} & \textbf{$D_{Test}$} & \textbf{$D_{Train}$} & \textbf{$D_{Test}$} & \textbf{$D_{Train}$} & \textbf{$D_{Test}$} & \textbf{$D_{Train}$} & \textbf{$D_{Test}$} \\
        \midrule
        \textbf{Fidelity} & \textbf{} & \textbf{} & \textbf{} & \textbf{} & \textbf{} & \textbf{} & \textbf{} & \textbf{} \\
        \midrule
        Column Shapes ($\uparrow$)& 0.777±0.004 & 0.778±0.008 & 0.914±0.002 & \textbf{0.913±0.004} & 0.844±0.018 & 0.839±0.021 & 0.899±0.009 & 0.885±0.011 \\
        Column Pair Trends ($\uparrow$) & 0.798±0.011 & 0.807±0.010 & 0.890±0.005 & \textbf{0.886±0.007} & 0.865±0.009 & 0.851±0.009 & 0.890±0.012 & 0.879±0.002 \\
        KSComplement ($\uparrow$) & 0.764±0.006 & 0.767±0.011 & 0.905±0.003 & \textbf{0.902±0.002} & 0.820±0.021 & 0.814±0.024 & 0.906±0.009 & 0.887±0.011 \\
        TVComplement ($\uparrow$) & 0.823±0.007 & 0.820±0.012 & 0.947±0.004 & \textbf{0.951±0.017} & 0.932±0.026 & 0.930±0.031 & 0.874±0.021 & 0.877±0.031 \\
        CorrelationSimilarity ($\uparrow$) & 0.947±0.005 & 0.947±0.005 & 0.987±0.002 & \textbf{0.969±0.004} & 0.913±0.006 & 0.917±0.007 & 0.961±0.006 & 0.946±0.009 \\
        ContingencySimilarity ($\uparrow$) & 0.740±0.011 & 0.739±0.017 & 0.919±0.005 & \textbf{0.916±0.019} & 0.893±0.037 & 0.885±0.043 & 0.795±0.036 & 0.801±0.044 \\
        WD ($\downarrow$) & 2.213±0.053 & 2.274±0.126 & 0.953±0.042 & 1.023±0.077 & 2.079±0.336 & 2.14±0.340 & 0.791±0.095 & \textbf{1.009±0.144}\\
        JSD ($\downarrow$) & 0.152±0.001 & \textbf{0.154±0.003} & 0.193±0.002 & 0.195±0.004 & 0.218±0.007 & 0.219±0.007 & 0.172±0.004 & 0.173±0.004 \\
        LogisticDetection (1 - ROC AUC) ($\uparrow$) & 0.380±0.015 & 0.398±0.022 & 0.830±0.032 & \textbf{0.866±0.040} & 0.398±0.112 & 0.405±0.118 & 0.566±0.066 & 0.553±0.089 \\
        \midrule
        \textbf{Privacy} & \textbf{} & \textbf{} & \textbf{} & \textbf{} & \textbf{} & \textbf{} & \textbf{} & \textbf{} \\
        \midrule
        NewRowSynthesis & 1±0 & 1±0 & 1±0 & 1±0 & 1±0 & 1±0 & 1±0 & 1±0 \\
        DCR (5th p.) ($\uparrow$) & \textbf{1.556±0.038} & 1.548±0.065 & 1.347±0.030 & 1.33±0.016 & 1.391±0.042 & 1.402±0.054 & 1.192±0.048 & 1.226±0.036 \\
        NNDR (5th p.) ($\uparrow$) & 0.765±0.012 & 0.769±0.021 & 0.727±0.016 & 0.707±0.015 & 0.737±0.015 & 0.738±0.018 & \textbf{0.768±0.016} & 0.778±0.021 \\
        CategoricalCAP ($\uparrow$) & 0.882±0.013 & 0.884±0.015 & \textbf{0.911±0.010} & 0.917±0.012 & 0.897±0.015 & 0.895±0.019 & 0.857±0.017 & 0.869±0.018 \\
        \midrule
        \textbf{Utility} & \textbf{} & \textbf{} & \textbf{} & \textbf{} & \textbf{} & \textbf{} & \textbf{} & \textbf{} \\
        \midrule
        TSTR (F1 score) ($\uparrow$)& - & \textbf{0.979±0.023} & - & 0.960±0.014 & - & 0.838±0.117 & - & 0.953±0.044 \\
        TATR (F1 score) ($\uparrow$)& - & \textbf{0.983±0.018} & - & 0.969±0.008 & - & 0.883±0.093 & - & 0.960±0.036 \\
        \bottomrule
    \end{tabular}
\end{table*}

\end{document}